\pdfoutput=1

\documentclass[sigconf,nonacm]{acmart}

\usepackage{algorithmic}
\usepackage{graphicx}
\usepackage{textcomp}
\usepackage{xcolor}

\usepackage{url}
\usepackage[T1]{fontenc}
\usepackage[utf8]{inputenc}
\usepackage{microtype}
\usepackage{tabularx}
\usepackage{booktabs}
\usepackage{multirow}

\DeclareGraphicsExtensions{.pdf,.ai,.jpg,.png}
\graphicspath{{.}}

\newcommand{\bsfigure}[3][scale=1.0]{%
  \begin{figure}[tb]
    \centering
    \includegraphics[#1]{#2}
    \vspace{-4ex}
    \caption{#3}\label{#2}
  \end{figure}}

\usepackage{xspace}
\newcommand{\corpusname}{SMAuC\xspace}

\newcommand{\pct}[1]{\textcolor{gray!75}{\small\xspace\textit{({#1}\%)}}}%

\newcommand{\Ni}{(1)~}
\newcommand{\Nii}{(2)~}
\newcommand{\Niii}{(3)~}
\newcommand{\Niv}{(4)~}
\newcommand{\Nv}{(5)~}

\begin{document}

\title{\corpusname\ -- The Scientific Multi-Authorship Corpus}

 \settopmatter{authorsperrow=4}

 \author{Janek Bevendorff}
 \affiliation{\normalsize
  \institution{Bauhaus-Universit\"at Weimar}
  \city{}
  \country{}
 }
 \orcid{0000-0002-3797-0559}

 \author{Philipp Sauer}
 \affiliation{\normalsize
  \institution{Leipzig University}
  \city{}
  \country{}
 }

 \author{Lukas Gienapp}
 \affiliation{\normalsize
  \institution{Leipzig University}
  \city{}
  \country{}
 }
 \orcid{0000-0001-5707-3751}

 \author{Wolfgang Kircheis}
 \affiliation{\normalsize
  \institution{Leipzig University}
  \city{}
  \country{}
 }
 \orcid{0000-0002-0925-0503}

 \author{Erik K\"orner}
 \affiliation{\normalsize
  \institution{Leipzig University}
  \city{}
  \country{}
 }
 \orcid{0000-0002-5639-6177}

 \author{Benno Stein}
 \affiliation{\normalsize
  \institution{Bauhaus-Universit\"at Weimar}
  \city{}
  \country{}
 }
 \orcid{0000-0001-9033-2217}

 \author{Martin Potthast}
 \affiliation{\normalsize
  \institution{Leipzig University}
  \city{}
  \country{}
 }
 \orcid{0000-0003-2451-0665}

\renewcommand{\shortauthors}{Bevendorff et al.}


\date{}

\begin{abstract}
The rapidly growing volume of scientific publications offers an interesting challenge for research on methods for analyzing the authorship of documents with one or more authors. However, most existing datasets lack scientific documents or the necessary metadata for constructing new experiments and test cases. We introduce \corpusname, a comprehensive, metadata-rich corpus tailored to scientific authorship analysis. Comprising over 3~million publications across various disciplines from over 5~million authors, \corpusname is the largest openly accessible corpus for this purpose. It encompasses scientific texts from humanities and natural sciences, accompanied by extensive, curated metadata, including unambiguous author~IDs. \corpusname aims to significantly advance the domain of authorship analysis in scientific texts.
\end{abstract}

\maketitle
\section{Introduction}

Authorship analysis focuses on distinguishing writing styles or attributing them to specific authors. Originating in the 19th century~\cite{koppel:2009}, this field has developed through various methods from linguistics, psychology, and notably, computer science. However, computational authorship analysis still struggles with scientific papers, as they tend to be relatively short and often contain a mix of co-authors' writing styles. Furthermore, many disciplines impose strict stylistic guidelines, limiting personal expression. Consequently, extracting individual stylistic traits from multi-authored documents remains a significant challenge.

To address the problem of multi-author authorship analysis, a comprehensive stylometric comparison of monographs and multi-author documents involving the same author is required. Identifying an author's stylistic features in multi-author documents allows for more accurate identification of their contributions. To facilitate corresponding advancements in computational authorship analysis for scientific texts, we introduce \corpusname, a dataset of 3,356,686~scientific papers with both single and multi-author origins, accompanied by detailed and disambiguated metadata. To our knowledge, this is the largest dataset compiled specifically for authorship analysis, suitable for research on scientific texts and broader stylometric investigations.%
\footnote{Code: \url{https://github.com/webis-de/JCDL-23}}%
$^,$%
\footnote{Download: \url{http://doi.org/10.5281/zenodo.7289788}}

We first examine current publicly accessible scientific text corpora (Section~\ref{sec:related}), then detail our dataset creation methodology (Section~\ref{sec:method}). Subsequently, we provide a qualitative and quantitative analysis of the dataset, its metadata, and the entire corpus (Section~\ref{sec:data}). Lastly, we discuss corpus applications (Section~\ref{sec:conclusion}), followed by and ethical self-assessment.

\section{Related Work}\label{sec:related}

While research on authorship has yielded several datasets in the scientific field, very few are available or can be reproduced: \citet{payer:2015} collect 6,872 conference papers in an effort to develop methods for de-anonymization of scientific publications. \citet{sarwar:2018} aggregate 2,573~papers from the arXiv preprint service to conduct multi-author attribution; similarly, \citet{rexha:2016} collect 6,144~articles from the PubMed database for the same purpose. \citet{boumber:2018} introduce and publicly release the MLPA-400 dataset, which consists of 400~scientific publications. Larger multi-author collections are available for other domains, for example the PAN-20 Style Change Detection Corpus~\cite{zangerle:2020} consisting of approximately 23,000~stack exchange postings, combining questions with answers to form multi-author documents.

Of the already very limited number of corpora that include large amounts of scientific texts, none were specifically designed for authorship analysis: \citet{soares:2019} use a self-constructed corpus of roughly 30,000 scientific documents in Portuguese, English, and Spanish for research on automated translation. \citet{citron:2015} present a corpus of 757,000 scientific texts for text reuse detection extracted from arXiv. \citet{gipp:2014} introduce a dataset of 234,591 articles extracted from the PubMed Central Open Access Subset, a large collection of biomedical full texts, many of which are available with an open access license. A corpus of 1.14~million paper full texts was used by both \citet{ammar:2018} and \citet{beltagy:2019}. The papers were obtained from Semantic Scholar and originate from computer science and biomedical research.

All of the corpora mentioned above exhibit one or several shortcomings: They are either very small and therefore only of limited use to large-scale authorship attribution, they lack the metadata required for authorship analyses, or they are too narrow in scope, i.e., limited to one scientific domain only. This necessitates the creation of a new large-scale dataset specifically curated for this purpose, which covers a larger variety of scientific disciplines with detailed metadata.

\section{Dataset Creation}\label{sec:method}

\corpusname is created by merging data from two sources: the CORE database~\cite{knoth:2011,knoth:2012}, a large collection of metadata and full texts of open access scientific publications, and the Microsoft Open Academic Graph (OAG,~\cite{sinha:2015}), a large, heterogeneous knowledge graph based on scientific articles, authors, and institutions.

As a basis for our dataset, we used the CORE%
\footnote{\url{https://core.ac.uk/services/dataset}}
database dump from 2018-03-01. It comprises 123M~metadata items, of which 85.6M items have abstracts and 9.8M~items have the full texts. Each item represents a single scientific paper or book. The OAG serves as an additional source for identifying and disambiguating the authors and fields of study of the publications. We rely on Version~2 of the OAG~\cite{hu:2020} with 179~million nodes and 2~billion edges.%
\footnote{\url{https://www.microsoft.com/en-us/research/project/open-academic-graph/}}

Table~\ref{table-core-filtering} illustrates the four-step selection process we applied to all entries in the final corpus:
\Ni
From the CORE corpus, we selected all entries with full texts and
\Nii
filtered these for English-language articles.
\Niii
We matched the selected subset with their corresponding OAG metadata to obtain unique author and fields of study information.
\Niv
Finally, we applied certain text quality heuristics for ensuring a high-quality extraction.

From the 123M CORE entries, we extracted a total of 9.8M entries with available full texts. Although CORE specifies a language flag, it is only present in some entries. We added missing language flags using a standard fastText~\cite{joulin:2016,joulin:2016b} language detection model. The texts were split into five parts of equal length of which at least four needed to be English. After this step, 6,531,442~entries remained.

\begin{table}[t]
\caption{Dataset curation process with number of documents remaining after each filtering step. Percentages relative to full CORE.}
\label{table-core-filtering}%
\centering
\renewcommand{\tabcolsep}{10pt}
\begin{tabular}{@{}lr@{\ \ }r@{}}
\toprule
  \textbf{Filtering applied}                     & \multicolumn{2}{c@{}}{\textbf{Number of documents}}                                                                     \\
\midrule
  CORE                                            &                                                                                       123,988,821 &           \pct{100.00} \\
  \quad $\hookrightarrow$ full texts              &                                                                                         9,835,064 & \pct{\phantom{00}7.93} \\
  \quad $\hookrightarrow$ text language English   &                                                                                         6,531,442 & \pct{\phantom{00}5.27} \\
  \quad $\hookrightarrow$ OAG matching            &                                                                                         3,508,509 & \pct{\phantom{00}2.82} \\
  \quad $\hookrightarrow$ text quality assurance  &                                                                                         3,356,686 & \pct{\phantom{00}2.70} \\
\bottomrule
\end{tabular}

\end{table}

\begin{table*}[t]
\caption{\textbf{(a)} Counts for all types of documents and their total; \textbf{(b)} Number of documents in the corpus by text length in characters and document type with percentage per row. Documents with no author information are omitted. Length values refer to the raw texts including tables, captions, and appendices.}
\label{table-document-types-and-lengths}%
\centering
\renewcommand{\tabcolsep}{16pt}
\begin{tabular}{@{}lr@{}}
  \multicolumn{2}{@{}l}{\textbf{(a)}}                 \\
\toprule
  \textbf{Document Type}             & \textbf{Count} \\
\midrule
  Single author without multi author &        711,471 \\
  Single author with multi author    &        261,629 \\
  Multi author without single author &      1,481,106 \\
  Multi author with single author    &        894,945 \\
  No author information              &          7,535 \\
\midrule
  Total                              &      3,356,686 \\
\bottomrule
\end{tabular}
\hfill%
\renewcommand{\tabcolsep}{12pt}
\begin{tabular}{@{}rrr@{\ \ }rr@{\ \ }r@{}}
  \multicolumn{6}{@{}l}{\textbf{(b)}}                                                                                       \\
\toprule
  \textbf{Length} & \textbf{Total} & \multicolumn{2}{c}{\textbf{Single author}} & \multicolumn{2}{c}{\textbf{Multi author}} \\
\midrule
     $\leq$ 3,000 &         39,300 &  13,680 &           \pct{\phantom{00}1.41} &    25,567 &        \pct{\phantom{00}1.07} \\
         -- 5,000 &         96,067 &  32,059 &           \pct{\phantom{00}3.29} &    63,832 &        \pct{\phantom{00}2.69} \\
        -- 50,000 &      2,273,246 & 467,844 &           \pct{\phantom{0}48.07} & 1,799,435 &        \pct{\phantom{0}75.73} \\
       -- 250,000 &        771,756 & 301,975 &           \pct{\phantom{0}31.03} &   468,473 &        \pct{\phantom{0}19.72} \\
      $>$ 250,000 &        176,317 & 157,542 &           \pct{\phantom{0}16.19} &    18,744 &        \pct{\phantom{00}0.79} \\
\midrule
            Total &      3,356,686 & 973,100 &                     \pct{100.00} & 2,376,051 &                  \pct{100.00} \\
\bottomrule
\end{tabular}
\end{table*}

In the third step, entries were merged with metadata in the~OAG. An official mapping between the two already exists (Version 2019-04-01\textsuperscript{1}), yet it contains only 655K of the 6.5M~English entries. Furthermore, the DOIs (as given in CORE and OAG) were not accurate in some cases. Using these DOIs as keys could result in false positive and false negative matching errors. To reduce the number of matching errors, we defined two extra matching criteria of which at least one had to be met to count as a match: 
\Ni 
The DOIs of both entries had to be identical and both titles had to have a Levenshtein distance of less than~10\% of the length of the shorter title.
\Nii 
The titles and years of publication had to be identical and at least one author name had to appear in both entries with a low Levenshtein distance as detailed above.

With this method, we were able to match OAG metadata for 3.5M CORE entries, a significant improvement over the official CORE-OAG-mapping. However, postprocessing the metadata was required in some cases. The fields of study given in the OAG per publication are of varying granularity (e.g., `humanities' as a whole vs. `chemical solid-state research' as a subfield of chemistry). To establish a standardized, hierarchical scheme, we manually mapped the annotated disciplines to the \textit{DFG Classification of Scientific Disciplines and Research Areas}~\cite{deutscheforschungsgemeinschaft:2021}. The mapping was carried out manually by three persons at very high agreement. Cases of disagreement were discussed internally and subsequently unified. 
The final three-level hierarchy includes {\em disciplines} (Engineering Sci., Humanities \& Social Sci., Life Sci., Natural Sci.), {\em research areas} (e.g. Chemistry, Medicine, Mechanical and Industrial Engineering,~\dots), and {\em fields} (e.g. Educational Research, Condensed Matter Physics, Zoology,~\dots).

In the final quality assurance step, the full texts of all entries were cleaned by removing markup and all non-ASCII characters, converting texts to lower case letters and collapsing runs of whitespace characters. Then, two heuristics were used to eliminate texts of sub-par quality: 
\Ni 
Cleaned texts with a length below 2,000~characters (approximately one printed page) were excluded.
\Nii 
Cleaned texts were split at sentence boundaries into three equally sized chunks and the fastText language detection model was once again applied to each part individually. If fastText considered a part to be English with more than 60\%~confidence, this part was accepted as English. An entry was excluded if more than one of the three parts was not classified as English. 
This repeated round of language classification was to further ensure that only English texts remain, since the first (coarse) round was performed on the uncleaned texts. The small number of entries~(152,000) removed in this step suggests that the coarse filtering already excluded non-English text reliably.

\section{Corpus Description}\label{sec:data}

This section describes the structure, format, and key properties of \corpusname. The corpus is distributed in the form of multiple line-delimited JSON files, each containing 100,000 entries. Each entry has identifiers (DOI, CORE, OAG) and detailed meta information about the publication (title, abstract, citation count, reference), its authors (name and OAG identifier), the discipline and field of study, and the full text from CORE.

\paragraph{Corpus Size and Composition}
Table~\ref{table-document-types-and-lengths}a details the composition of the corpus itemized by document type. Publications can be split into two fundamental categories:
\Ni
single-author (i.e., monographs) and
\Nii
multi-author (i.e., collaborative) publications. By investigating author relations, we can further differentiate each of the two into sub-types, for a total of four document types:
\Ni 
single-author publications whose authors have not participated in any multi-author publications;
\Nii
single-author publications whose authors appear in at least one multi-author publication;
\Niii 
multi-author publications whose authors have not written any monographs; and 
\Niv
multi-author publications with at least one author who has written at least one additional monograph.
In addition, for a very small subset of documents, no author information is available. These texts will not be immediately useful for attributing the texts to specific authors, though they may still be useful material if larger collections of text from specific research areas are needed, e.g., for comparisons between sub-corpora of the humanities and sciences.

The corpus contains fewer monographs than multi-author documents. Only a minority of monograph authors have participated in multi-author publications and vice versa. Documents with no author information are rare. The total document count exceeds previous datasets on authorship analysis.

\paragraph{Text Lengths} 
The corpus comprises a wide range of texts of different lengths: from very short articles of just a few pages to long book-sized entries. Table~\ref{table-document-types-and-lengths}b lists document counts for different length bins. The bins were chosen as approximate character counts for \Ni abstract papers (less than one page), \Nii short papers (one or two pages), \Niii essay-length papers (up to 10~pages), \Niv long papers (up to 50~pages), and \Nv books or dissertations (more than 50~pages). Most papers in the corpus are between 5,000 and 50,000 characters in length (2~and 20~pages). Multi-author publications are shorter on average than single-author publications. A sizable portion of publications exceed lengths of 250,000 characters, most of which are monographs. These seem to be mainly individual dissertations and less often collaborative book publications. Manual spot checks confirmed this assumption.

\begin{table}[t]
\caption{Single-author and multi-author documents, median authors per document (\textbf{Med.}) and median text length, by research area.}
\label{table-documents-by-discipline}
\centering
\renewcommand{\tabcolsep}{1pt}
\begin{tabular}{@{}lcccr@{}}
\toprule
  \textbf{Research Area} & \multicolumn{1}{c}{\textbf{Single author}} & \multicolumn{1}{c}{\textbf{Multi author}} & \multicolumn{1}{c}{\textbf{Med.}} & \multicolumn{1}{c@{}}{\textbf{Length}} \\
\midrule
  Engineering Sciences   &             \phantom{0}55,015              &                  375,206                  &                 3                 &                                 28,467 \\
  Humanities             &             \phantom{0}58,317              &                  199,926                  &                 3                 &                                 37,224 \\
  Life Sciences          &             \phantom{0}48,723              &                  715,218                  &                 5                 &                                 32,616 \\
  Natural Sciences       &                  147,024                   &                  651,076                  &                 3                 &                                 26,103 \\
\bottomrule
\end{tabular}
\end{table}

\paragraph{Academic Disciplines}
Fields of study annotations are available for approximately 1.7M entries in the corpus. Table~\ref{table-documents-by-discipline} lists document counts and key statistics per discipline, reflecting different publishing practices across fields. For example, the median text length is higher in the humanities compared to the natural sciences, while the median author count is highest for the life sciences. The relative proportion of single- and multi-author documents also differs per discipline, yet in all disciplines, sufficient amounts of either type are present to conduct authorship analyses.

\bsfigure[width=\linewidth]{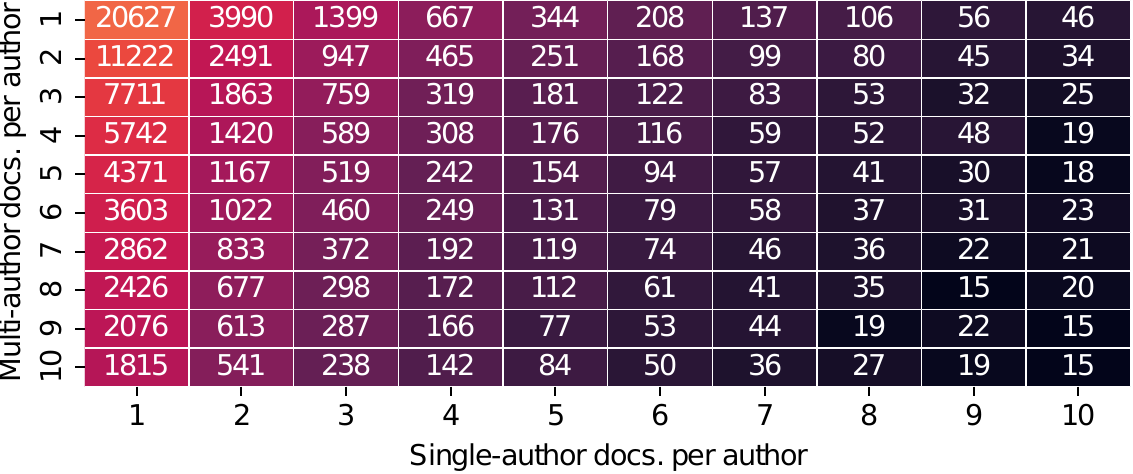}{Author count over number of single-author and multi-author publications per author. Counts beyond 10~are omitted (35,178 authors).\Description[Author count over number of single-author and multi-author publications per author. Counts beyond 10~are omitted (35,178 authors)]{Author count over number of single-author and multi-author publications per author.}}

\paragraph{Author Information}
Establishing reliable author relations between documents is paramount for use as a ground-truth for authorship analysis. To this end, the OAG provides disambiguated and unique author~IDs. Of particular interest to us are authors involved in both single- and multi-author publications. In total, 5,664,224 unique authors are present in the corpus. Of these, 670,566 appear exclusively in single-author publications and 4,868,263 exclusively in multi-author documents. The remaining 125,395 authors, who appear in both types of documents, are thus of particular interest. Figure~\ref{author-heatmap} shows author counts over the number of single-author and multi-author publications. Unsurprisingly, the vast majority of authors appear in at most one or two single- or multi-author publications, respectively. More than five documents per author in either category are increasingly rare.

The 973,100~monographs in the corpus can be attributed to 795,000~different authors. The majority of them is represented with just one monograph (92.02\%). This also means that 8\% of the authors have written more than 24\% of the monographs in the corpus. For those 8\% it will be possible to extract individual stylistic features based on a larger set of texts.
For a total of 125,395~authors, both single-author and multi-author documents are present, amounting to a total of 1.15M~documents. Most of these authors have written a small number of single-author and an even smaller number of multi-author documents. 

\paragraph{Corpus File Description} 
The corpus is distributed as \texttt{xz}-com\-pressed file in \texttt{JSONL} format, where records are each encoded as a single \texttt{JSON} string per line. The corpus can thus be efficiently stream-processed without requiring to inflate the corpus file. An example record with associated keys, datatypes, description is specified in Table~\ref{tab:fields}.

\begin{center}
\begin{table*}[t]
\caption{Schema for a single data record with field names, datatypes, descriptions, and randomly drawn example. }
\label{tab:fields}
\begin{tabular}{@{}l@{\hspace{.75em}}llll@{}}
\toprule
&\textbf{Key}                & \textbf{Datatype}                 & \textbf{Description}                          & \textbf{Example} \\
\midrule
\multirow{4}{*}{\rotatebox{90}{IDs}}
&\texttt{core\_id}           & \texttt{int}                      & ID of the publication in the CORE corpus      & 2461603 \\
&\texttt{doi}                & \texttt{string}                   & DOI of the publication                        & 10.1103/PhysRevD.63.123512 \\
&\texttt{download\_url}      & \texttt{string}                   & URL of fulltext PDF                           & \url{http://arxiv.org/abs/hep-ph/0012097} \\
&\texttt{mag\_ids}           & \texttt{array of int}             & Microsoft Academic Graph ID                   & [2064413872]    \\
\midrule
\multirow{9}{*}{\rotatebox{90}{Metadata}}
&\texttt{authors}            & \texttt{array of (int, string)}   & Author list with ID and name per author       & [(2238602696, Arttu Rajantie), ...] \\
&\texttt{doc\_type}          & \texttt{string}                   & Type of the publication                       & Journal \\
&\texttt{fields\_of\_study}  & \texttt{array of string}          & Array of field names                          & [Particles, Nuclei and Fields, ...] \\
&\texttt{publisher}          & \texttt{string}                   & Publisher of the publication                  & The American Physical Society  \\
&\texttt{venue}              & \texttt{(int, string) tuple}      & Venue ID and name                             & (173952182, Physics Letters B) \\
&\texttt{issue}              & \texttt{string}                   & Issue the publication appeared in             & 12 \\
&\texttt{volume}             & \texttt{string}                   & Volume the publication appeared in            & 63 \\
&\texttt{year}               & \texttt{int}                      & Year of publication                           & 2001 \\
&\texttt{n\_citation}        & \texttt{int}                      & Number of citations                           & 29 \\
\midrule
\multirow{3}{*}{\rotatebox{90}{Text}}
&\texttt{title}              & \texttt{string}                   & Title of the publication                     & Electroweak preheating on a lattice \\
&\texttt{abstract}           & \texttt{string}                   & Abstract text                                 & In many inflationary models, large ...\\
&\texttt{fulltext}           & \texttt{string}                   & Parsed fulltext                               & In many inflationary models, large ... \\
\bottomrule
\end{tabular}
\end{table*}
\end{center}
\section{Conclusion}\label{sec:conclusion}

We introduce \corpusname, the largest available corpus for authorship analysis in the scientific domain. It encompasses over 3.3M documents and detailed, standardized metadata including author and field-of-study annotations. The corpus allows to select subsets of texts according to numerous criteria, each in itself still met by a significant number of documents. Selecting only very short texts is just as possible as picking entire volumes; including only authors with a high number of monographs will still generate subsets with several thousands of texts. Even selecting only multi-author texts for which individual writing style analyses are supported by additional monographs, leaves a subset of more than 70,000~documents. If smaller subsets are sufficient, it is also possible to combine constraints, e.g. select all multi-author texts only from the humanities with additional monographs for all authors. The corpus allows for compiling relevant subsets tailored to very specific research questions in authorship analysis, particularly in, but not restricted to, scientific texts.

The core element of the corpus are the 1,144,915 documents for which monographies and multiauthor documents from the same authors are available, which could be of particular interest for authorship analysis in the context of multiauthorship. Future research with the corpus could include the application of well-tested authorship analysis methods like Unmasking \cite{koppel:2007} on the domain of academic texts. Those methods can now also be tested on their ability to detect writing styles of authors extracted from their monographies in documents co-authored by other researchers.

%

\section*{Ethics Statement}\label{sec:ethics}

Our dataset compiles contemporary writing from the domain of science (``papers'') with the purpose of studying the capabilities of authorship analysis technology in dealing with scientific papers and the challenges that arise from multi-author documents. Ethical considerations for datasets in general relate to four main areas of concern \cite{peng:2021}, three of which are relevant to this paper:
\Ni
privacy of the individuals included in the data, 
\Nii
effects of biases on downstream use, and
\Niii
dataset usage for dubious purposes. We therefore took into account a consensus on best-practices for ethical dataset creation \cite{gebru:2021,leidner:2017,mieskes:2017}.

Ad~(1).
An anonymization or pseudonymization of the papers in our corpus is virtually impossible, since they are publicly available, and querying for the original CORE/OAG data would reveal the author(s) of every enclosed paper. By partaking in the scientific discourse, however, any published paper becomes part of science's legacy, which is open to everyone to make it their subject of analysis, scrutiny, and mining. This is especially true for articles under an open-access license, where consent to the creation of derivative works, public archiving, and mining is implied.

Ad~(2).
Stylometry is particularly prone to confounding variables such as text domain, genre, or audience \cite{bevendorff:2019,bischoff:2020,koolen:2017}, which replicates to downstream tasks. No explicit measures for preventing such biases in the data can be taken given the wide variety of authorship-related tasks that can be studied. Rather, we opt to include as much data and metadata as possible to enable researchers to derive their own datasets for their specific tasks, allowing them to address confounding factors individually. The dataset strives for transparency and extendability by documenting its creation process and by retaining references to the original data sources.
It is a collection of texts and metadata obtained from publicly available sources, assembled with respect to their terms and conditions. 

Ad~(3).
We deem the overall abuse potential of the corpus low, particularly in comparison to what is already possible today with the~OAG. Yet, as a further precaution, access to the data will be granted on a per-request basis via Zenodo for academic use only.

\newpage


\begin{raggedright}
\bibliographystyle{ACM-Reference-Format}
\bibliography{jcdl23-multi-authorship-corpus-lit-shortened}
\end{raggedright}

\end{document}